\documentclass[runningheads]{llncs}

\usepackage{graphicx}
\usepackage[colorinlistoftodos]{todonotes}
\usepackage{subfig}
\usepackage{url}

\begin{document}

\title{Data augmentation using synthetic data for time series classification with deep residual networks}

\titlerunning{Data augmentation using synthetic data for TSC with deep ResNets}

\author{Hassan Ismail Fawaz, 
Germain Forestier, 
Jonathan Weber, \\
Lhassane Idoumghar \and
Pierre-Alain Muller
}

\authorrunning{Ismail Fawaz et al.}

\institute{
IRIMAS, Universit\'{e} de Haute-Alsace, 68100 Mulhouse, France\\
\email{\{first-name\}.\{last-name\}@uha.fr}
}

\maketitle            

\begin{abstract}
Data augmentation in deep neural networks is the process of generating artificial data in order to reduce the variance of the classifier with the goal to reduce the number of errors. 
This idea has been shown to improve deep neural network's generalization capabilities in many computer vision tasks such as image recognition and object localization. 
Apart from these applications, deep Convolutional Neural Networks (CNNs) have also recently gained popularity in the Time Series Classification (TSC) community. 
However, unlike in image recognition problems, data augmentation techniques have not yet been investigated thoroughly for the TSC task. 
This is surprising as the accuracy of deep learning models for TSC could potentially be improved, especially for small datasets that exhibit overfitting, when a data augmentation method is adopted. 
In this paper, we fill this gap by investigating the application of a recently proposed data augmentation technique based on the Dynamic Time Warping distance, for a deep learning model for TSC.
To evaluate the potential of augmenting the training set, we performed extensive experiments using the UCR TSC benchmark. 
Our preliminary experiments reveal that data augmentation can drastically increase deep CNN's accuracy on some datasets and significantly improve the deep model's accuracy when the method is used in an ensemble approach.
\end{abstract}

\section{Introduction}\label{sec-introduction}

Deep learning usually benefits from large training sets~\cite{zhang2017understanding}. 
However, for many applications only relatively small training data exist. 
In Time Series Classification (TSC), this phenomenon can be observed by analyzing the UCR archive's datasets~\cite{ucrarchive}, where 20 datasets have 50 or fewer training instances. 
These numbers are relatively small compared to the billions of labeled images in computer vision, where deep learning has seen its most successful applications~\cite{lecun2015deep}.

\begin{figure}
\centering
    \subfloat[DiatomSizeReduction]{
 \includegraphics[width=0.47\linewidth]{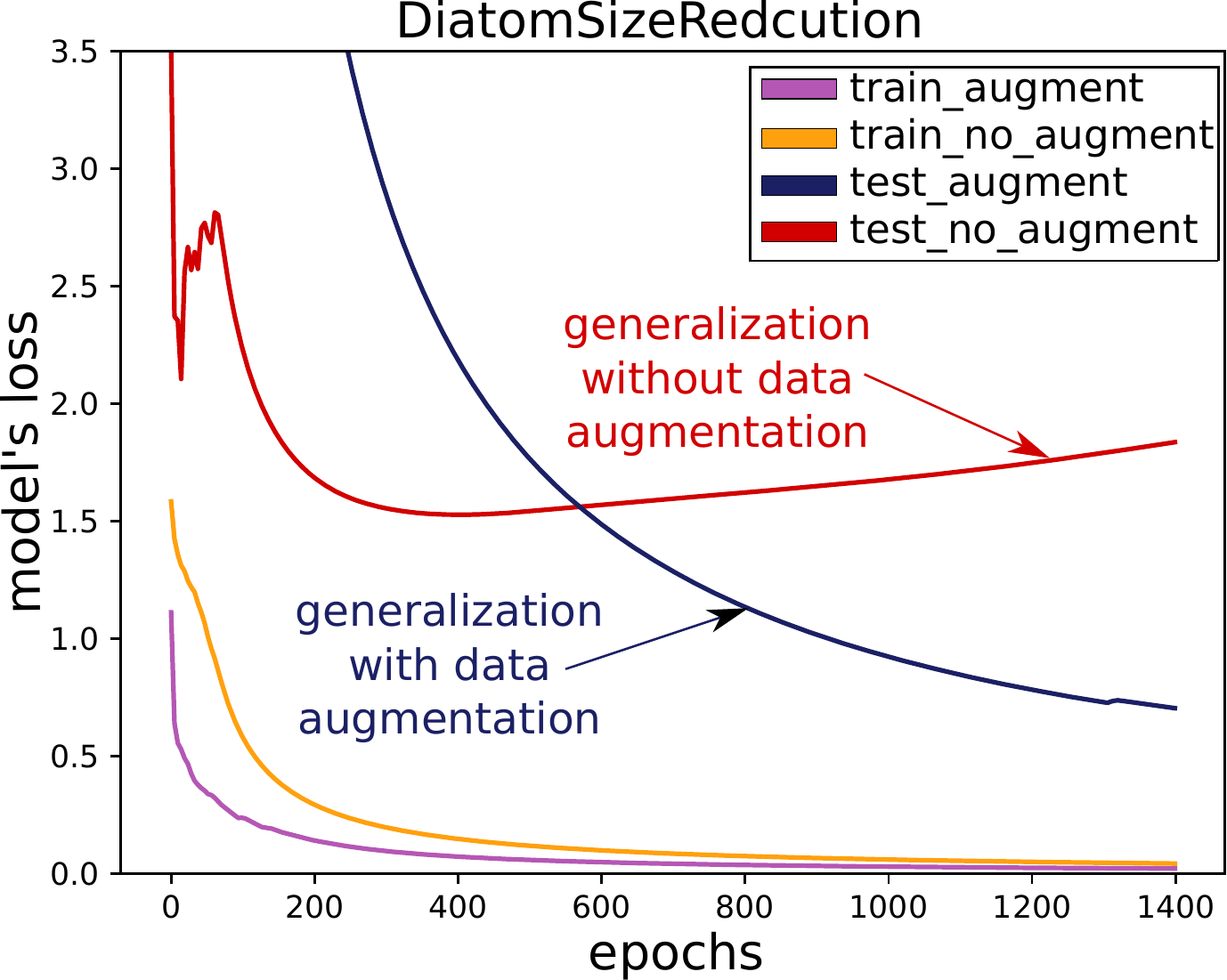}
      \label{sub-diatomsizereduction}}
 \hspace{.1cm} 
    \subfloat[Meat]{
 \includegraphics[width=0.47\linewidth]{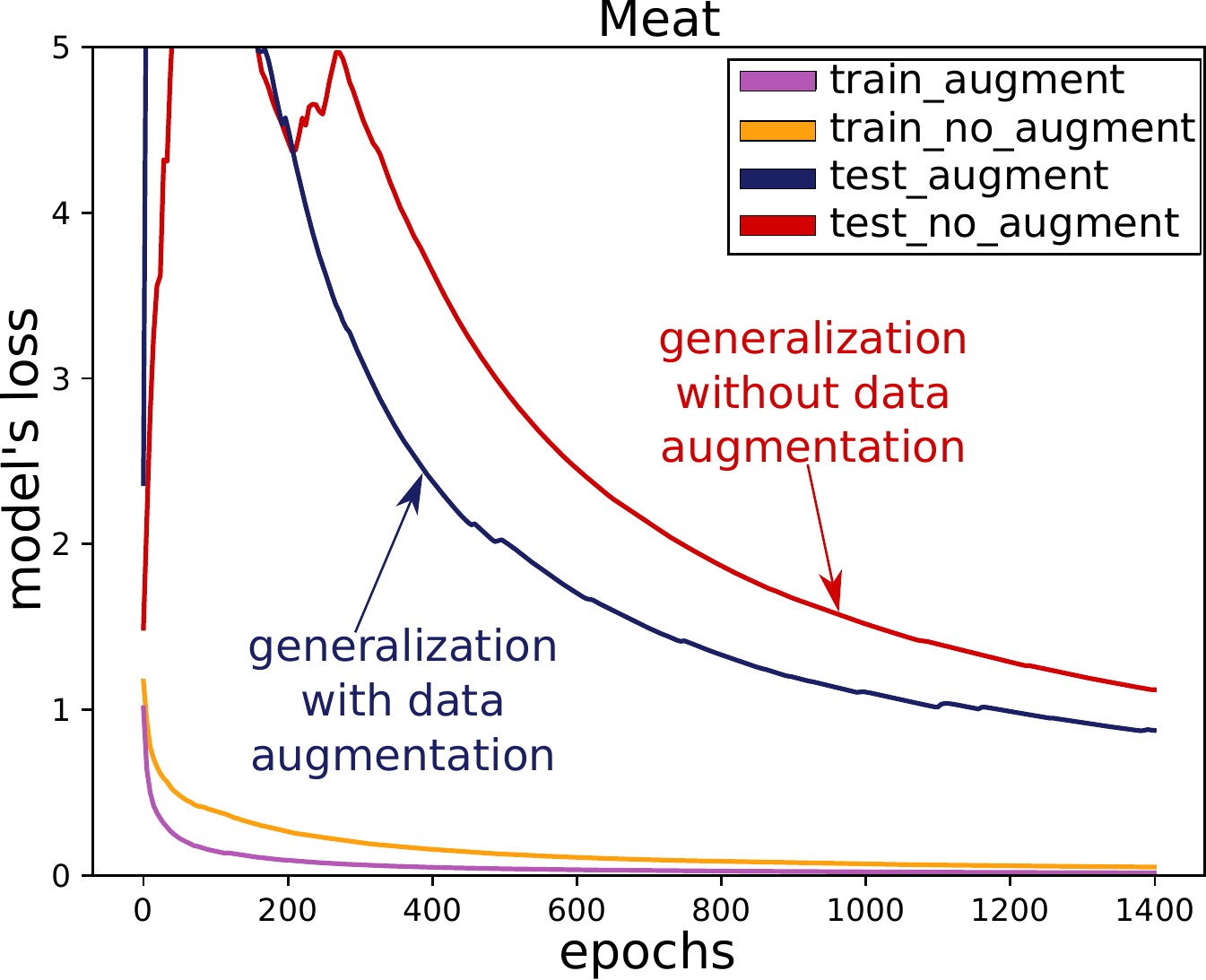}
      \label{sub-meat}}
    \caption{The model's loss with/without data augmentation on the DiatomSizeReduction and Meat datasets (smoothed and clipped for visual clarity).}
    \label{fig-plot-generalization}
\end{figure}

Although the recently proposed deep Convolutional Neural Networks (CNNs) reached state of the art performance in TSC on the UCR archive~\cite{wang2017time}, they still show low generalization capabilities on some small datasets such as the DiatomSizeReduction dataset. 
This is surprising since the nearest neighbor approach (1-NN) coupled with the Dynamic Time Warping (DTW) performs exceptionally well on this dataset which shows the relative easiness of this classification task. 
Thus, inter-time series similarities in such small datasets cannot be captured by the CNNs due to the lack of labeled instances, which pushes the network's optimization algorithm to be stuck in local minimums~\cite{zhang2017understanding}. 
\figurename~\ref{fig-plot-generalization} illustrates on an example that the lack of labeled data can sometimes be compensated by the addition of synthetic data.

This phenomenon, also known as \emph{overfitting} in the machine learning community, can be solved using different techniques such as regularization or simply collecting more \emph{labeled} data~\cite{zhang2017understanding} (which in some domains are hard to obtain).  
Another well-known technique is data augmentation, where synthetic data are generated using a specific method. 
For example, images containing street numbers on houses can be slightly rotated without changing what number they actually are~\cite{krizhevsky2012imagenet}.  
For deep learning models, these methods are usually proposed for image data and do not generalize well to time series~\cite{um2017data}.
This is probably due to the fact that for images, a visual comparison can confirm if the transformation (such as rotation) did not alter the image's class, while for time series data, one cannot easily confirm the effect of such ad-hoc transformations on the nature of a time series.  
This is the main reason why data augmentation for TSC have been limited to mainly two relatively simple techniques: slicing and manual warping, which are further discussed in Section~\ref{sec-related}.   

In this paper, we propose to leverage from a DTW based data augmentation technique specifically developed for time series, in order to boost the performance of a deep Residual Network (ResNet) for TSC.
Our preliminary experiments reveal that data augmentation can drastically increase the accuracy for CNNs on some datasets while having a small negative impact on other datasets.
We finally propose to combine the decision of the two trained models and show how it can reduce significantly the rare negative effect of data augmentation while maintaining its high gain in accuracy on other datasets.

\section{Related work}\label{sec-related}
The most used data augmentation method for TSC is the slicing window technique, originally introduced for deep CNNs in~\cite{cui2016multi}. 
The method was originally inspired by the image cropping technique for data augmentation in computer vision tasks~\cite{zhang2017understanding}. 
This data transformation technique can, to a certain degree, guarantee that the cropped image still holds the same information as the original image. 
On the other hand, for time series data, one cannot make sure that the discriminative information has not been lost when a certain region of the time series is cropped. 
Nevertheless, this method was used in several TSC problems, such as in~\cite{krell2018data} where it improved the Support Vector Machines accuracy for classifying electroencephalographic time series. 
In~\cite{kvamme2018predicting}, this slicing window technique was also adopted to improve the CNNs' mortgage delinquency prediction using customers' historical transactional data. 
In addition to the slicing window technique, jittering, scaling, warping and permutation were proposed in~\cite{um2017data} as generic time series data augmentation approaches.
The authors in~\cite{um2017data} proposed a novel data augmentation method specific to wearable sensor time series data that rotates the trajectory of a person's arm around an axis (e.g. $x$ axis). 

In~\cite{leguennec2016data}, the authors proposed to extend the slicing window technique with a warping window that generates synthetic time series by warping the data through time. 
This method was used to improve the classification of their deep CNN for TSC, which was also shown to significantly decrease the accuracy of a NN-DTW classifier when compared to our adopted data augmentation algorithm~\cite{forestier2017generating}.
We should note that the use of a window slicing technique means that the model should classify each subsequence alone and then finally classify the whole time series using a majority voting approach. 
Alternatively, our method does not crop time series into shorter subsequences which enables the network to learn discriminative properties from the whole time series in an end-to-end manner.

\section{Method}\label{sec-method}

\subsection{Architecture}

\begin{figure}
\includegraphics[]{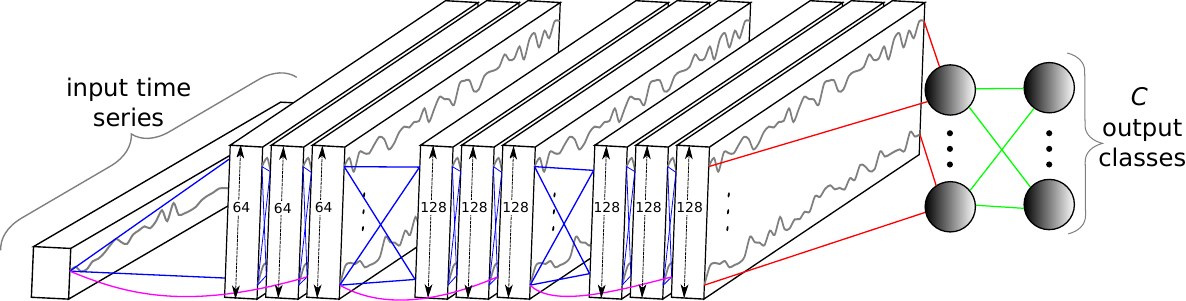}
\caption{Residual network's architecture. 
Blue connections correspond to convolutions, violet to residual, red to average pooling and green are fully-connected.}\label{fig-resnet-archi}
\end{figure}

We have chosen to improve the generalization capability of the deep ResNet proposed in~\cite{wang2017time} for two main reasons, whose corresponding architecture is illustrated in \figurename~\ref{fig-resnet-archi}.  
First, by adopting an already validated architecture, we can attribute any improvement in the network's performance solely to the data augmentation technique. 
The second reason is that ResNet~\cite{wang2017time}, to the best of our knowledge, is the deepest neural network validated on large number of TSC tasks (such as the UCR archive~\cite{ucrarchive}), which according to the deep learning literature will benefit the most from the data augmentation techniques as opposed to shallow architectures~\cite{bengio2011deep}.
Deep ResNets were first proposed by He et al.~\cite{he2016deep} for computer vision tasks.
They are mainly composed of convolutions, with one important characteristic: the residual connections which acts like shortcuts that enable the flow of the gradient directly through these connections. 

The input of this network is a univariate time series with a varying length $l$. 
The output consists of a probability distribution over the $C$ classes in the dataset. 
The network's core contains three residual blocks followed by a Global Average Pooling layer and a final softmax classifier with $C$ neurons. 
Each residual block contains three 1-D convolutions of respectively 8, 5 and 3 filter lengths.
Each convolution is followed by a batch normalization~\cite{ioffe2015batch} and a Rectified Linear Unit (ReLU) as the activation function.
The residual connection consists in linking the input of a residual block to the input of its consecutive layer with the simple addition operation. 
The number of filters in the first residual blocks is set to 64 filters, while the second and third blocks contains 128 filters.  

All network's parameters were initialized using Glorot's Uniform initialization method~\cite{glorot2010understanding}. 
These parameters were learned using Adam~\cite{kingma2015adam} as the optimization algorithm. 
Following~\cite{wang2017time}, without any fine-tuning, the learning rate was set to $0.001$ and the exponential decay rates of the first and second moment estimates were set to $0.9$ and $0.999$ respectively.
Finally, the categorical cross-entropy was used as the objective cost function during the optimization process.  

\subsection{Data augmentation}

The data augmentation method we have chosen to test with this deep architecture, was first proposed in~\cite{forestier2017generating} to augment the training set for a 1-NN coupled with the DTW distance in a cold start simulation problem.
In addition, the 1-NN was shown to sometimes benefit from augmenting the size of the train set even when the whole dataset is available for training.
Thus, we hypothesize that this synthetic time series generation method should improve deep neural network's performance, especially that the generated examples in~\cite{forestier2017generating} were shown to closely follow the distribution from which the original dataset was sampled. 
The method is mainly based on a weighted form of DTW Barycentric Averaging (DBA) technique~\cite{petitjean2011a,petitjean2016faster,petitjean2014dynamic}. 
The latter algorithm averages a set of time series in a DTW induced space and by leveraging a weighted version of DBA, the method can thus create an infinite number of new time series from a given set of time series by simply varying these weights.
Three techniques were proposed to select these weights, from which we chose only one in our approach for the sake of simplicity, although we consider evaluating other techniques in our future work.
The weighting method is called Average Selected which consists of selecting a subset of close time series and fill their bounding boxes. 

We start by describing in details how the weights are assigned, which constitutes the main difference between an original version of DBA and the weighted version originally proposed in~\cite{forestier2017generating}. 
Starting with a random initial time series chosen from the training set, we assign it a weight equal to $0.5$. 
The latter randomly selected time series will act as the initialization of DBA.
Then, we search for its 5 nearest neighbors using the DTW distance. 
We then randomly select 2 out these 5 neighbors and assign them a weight value equal to 0.15 each, making thus the total sum of assigned weights till now equal to $0.5+2\times0.15=0.8$.
Therefore, in order to have a normalized sum of weights (equal to 1), the rest of the time series in the subset will share the rest of the weight $0.2$. 
We should note that the process of generating synthetic time series leveraged only the training set thus eliminating any bias due to having seen the test set's distribution. 

As for computing the average sequence, we adopted the DBA algorithm in our data augmentation framework.
Although other time series averaging methods exist in the literature, we chose the weighted version of DBA since it was already proposed as a data augmentation technique to solve the cold start problem when using a nearest neighbor classifier~\cite{forestier2017generating}.
Therefore we emphasize that other weighted averaging methods such as soft-DTW~\cite{cutur2017soft} could be used instead of DBA in our framework, but we leave such exploration for our future work. 

We did not test the effect of imbalanced classes in the training set and how it could affect the model's generalization capabilities.  
Note that imbalanced time series classification is a recent active area of research that merits an empirical study of its own~\cite{geng2018cost}.  
At last, we should add that the number of generated time series in our framework was chosen to be equal to double the amount of time series in the most represented class (which is a hyper-parameter of our approach that we aim to further investigate in our future work).   

\section{Results}\label{sec-results}

\subsection{Experimental Setup}

We evaluated the data augmentation method for ResNet on the UCR archive~\cite{ucrarchive}, which is the largest publicly available TSC benchmark.
The archive is composed of datasets from different real world applications with varying characteristics such the number of classes and the size of the training set. 
Finally, for training the deep learning models, we leveraged the high computational power of more than 60 GPUs in one huge cluster\footnote{Our source code is available on \url{https://github.com/hfawaz/aaltd18}}
We should also note that the same parameters' initial values were used for all compared approaches, thus eliminating any bias due to the random initialization of the network's weights.

\subsection{Effect of data augmentation}
\begin{figure}
\centering
    \subfloat[ResNet with/without augmentation]{
    
 \includegraphics[width=0.47\linewidth]{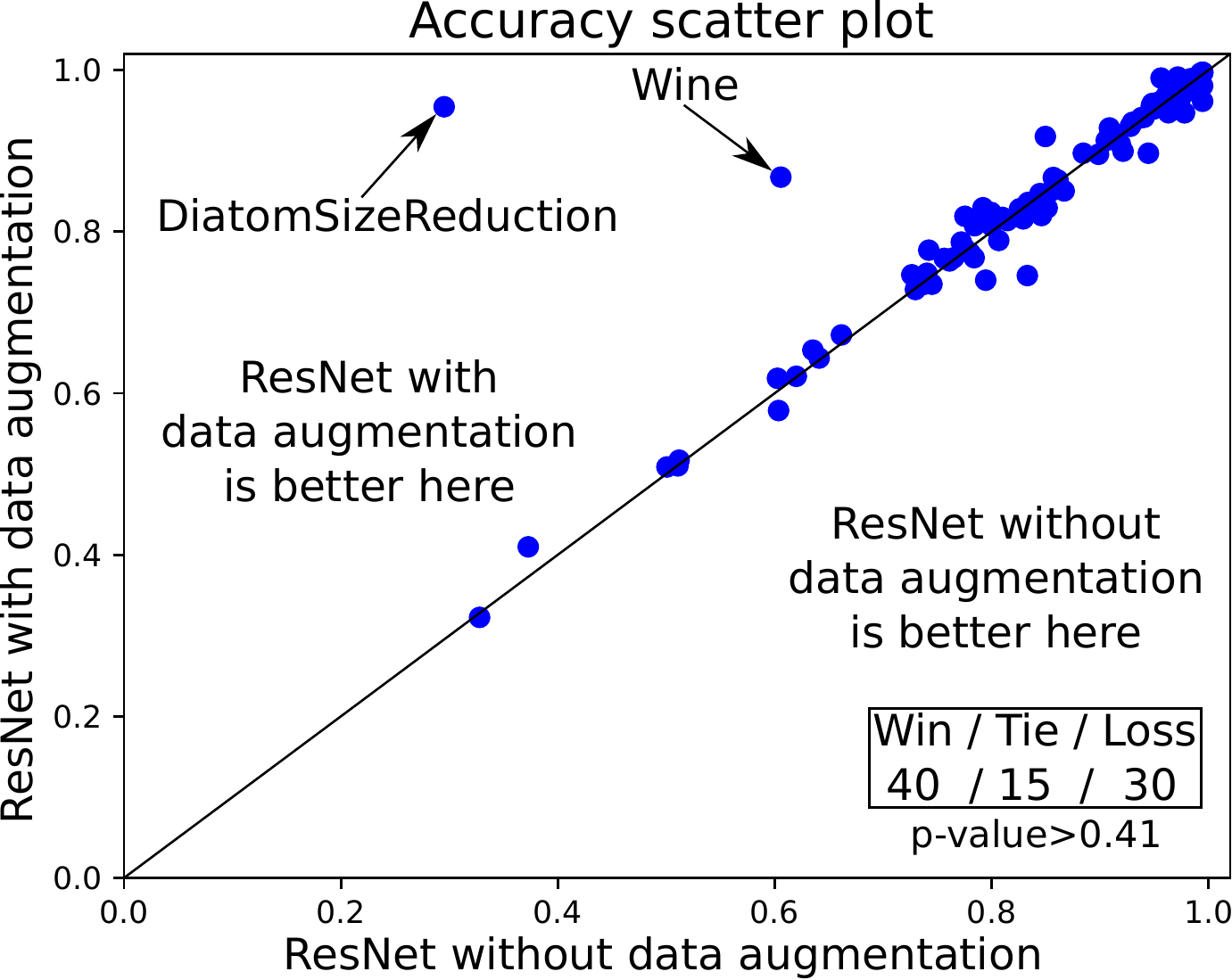}
      \label{sub-data-augment}}
 \hspace{.1cm}
    \subfloat[ResNet ensemble vs ResNet]{ 
 \includegraphics[width=0.47\linewidth]{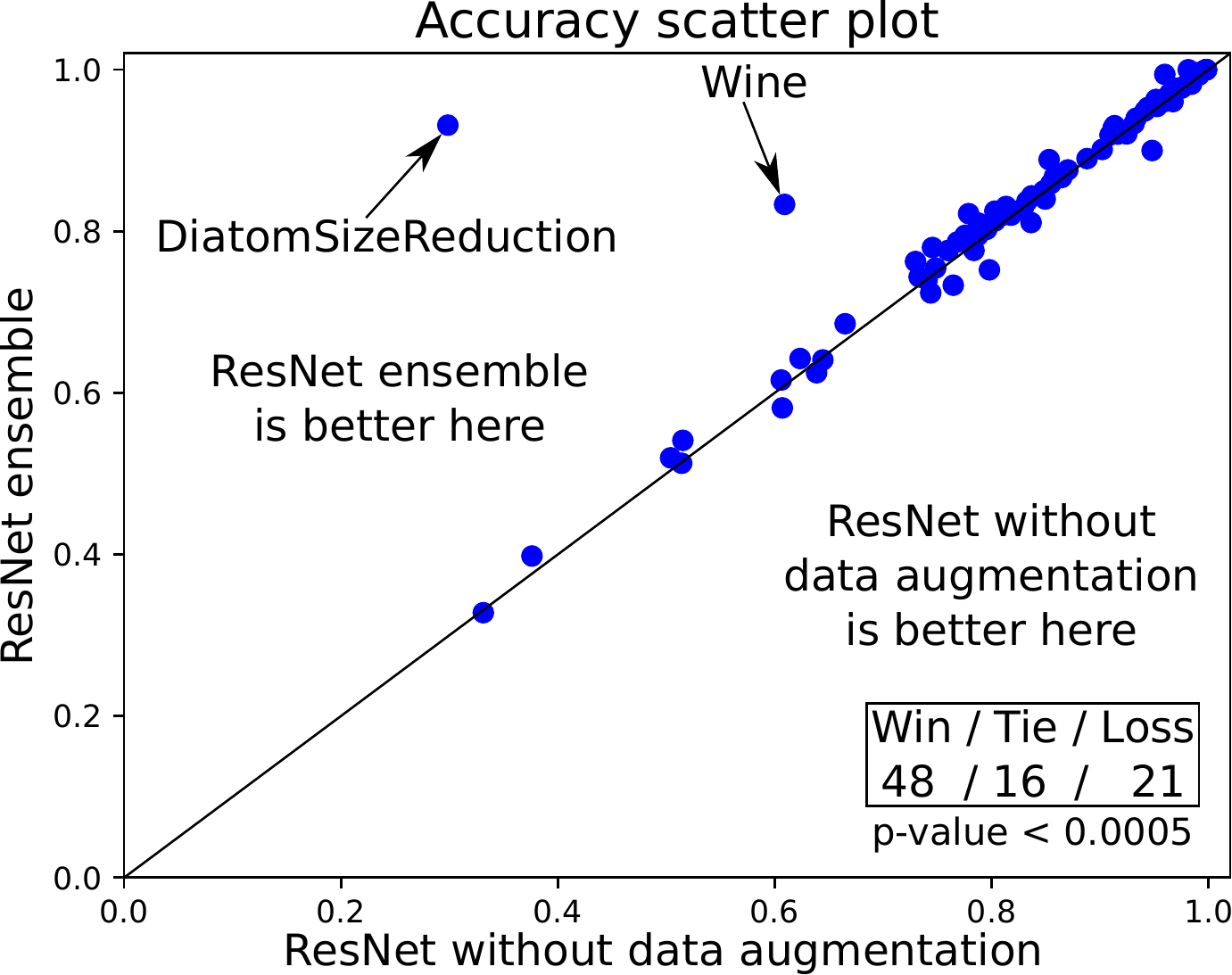}
      \label{sub-ensemble}
      }
    \caption{Accuracy of ResNet with and/or without data augmentation.}
    \label{fig-scatters}
\end{figure}

Our results show that data augmentation can drastically improve the accuracy of a deep learning model while having a small negative impact on some datasets in the worst case scenario.
\figurename~\ref{sub-data-augment} shows the difference in accuracy between ResNet with and without data augmentation, it shows that the data augmentation technique does not lead a significant decrease in accuracy. 
Additionally, we observe a huge increase of accuracy for the DiatomSizeReduction dataset (the accuracy increases from $30\%$ to $96\%$ when using data augmentation). 

This result is very interesting for two main reasons. 
First, DiatomSizeReduction has the smallest training set in the UCR archive~\cite{ucrarchive} (with 16 training instances), which shows the benefit of increasing the number of training instances by generating synthetic time series. 
Secondly, the DiatomSizeReduction dataset is the one where ResNet yield the worst accuracy without augmentation. 
On the other hand, the 1-NN coupled with DTW (or the Euclidean distance) gives an accuracy of $97\%$ which shows the relative easiness of this dataset where time series exhibit similarities that can be captured by the simple Euclidean distance, but missed by the deep ResNet due to the lack of training data (which is compensated by our data augmentation technique). 
The results for the Wine dataset (57 training instances) also show an important improvement when using data augmentation.  

While we did show that deep ResNet can benefit from synthetic time series on some datasets, we did not manage to show any significant improvement over the whole UCR archive ($p$-value $> 0.41$ for the Wilcoxon signed rank test).
Therefore, we decided to leverage an ensemble technique where we take into consideration the decisions of two ResNets (trained with and without data augmentation).
In fact, we average the a posteriori probability for each class over both classifier outputs, then assign for each time series the label for which the averaged probability is maximum, thus giving a more robust approach to out-of-sample generated time series. 
The results in \figurename~\ref{sub-ensemble} show that the datasets which benefited the most from data augmentation exhibit almost no change to their accuracy improvement. 
While on the other hand the number of datasets where data augmentation harmed the model's accuracy decreased from $30$ to $21$.  
The Wilcoxon signed rank test shows a significant difference ($p$-value $< 0.0005$).
The ensemble's results are in compliance with the recent consensus in the TSC community, where ensembles tend to improve the individual classifiers' accuracy~\cite{bagnall2017the}. 

\section{Conclusion}\label{sec-conclusion}
In this paper, we showed how overfitting small time series datasets can be mitigated using a recent data augmentation technique that is based on DTW and a weighted version of the DBA algorithm. 
These findings are very interesting since no previous observation made a link between the space induced by the classic DTW and the features learned by the CNNs, whereas our experiments showed that by providing enough time series, CNNs are able to learn time invariant features that are useful for classification.  

In our future work, we aim to further test other variant weighting schemes for the DTW-based data augmentation technique, while providing a method that predicts when and for which datasets, data augmentation would be beneficial.   

\subsubsection*{\ackname}\label{sec-acknowledgment}
We would like to thank NVIDIA Corp. for the Quadro P6000 grant and the M\'esocentre of Strasbourg for providing access to the cluster. 

\bibliographystyle{splncs04}
\bibliography{biblio}

\end{document}